\documentclass[nohyperref]{article}

\usepackage{microtype}
\usepackage{graphicx}
\usepackage{titlesec}
\usepackage{subfigure}
\usepackage{bm}
\usepackage{booktabs} 

\usepackage{hyperref}

\usepackage[accepted]{icml2022}

\usepackage{amsmath}
\usepackage{amssymb}
\usepackage{mathtools}
\usepackage{amsthm}
\usepackage{multirow}

\usepackage[capitalize,noabbrev]{cleveref}

\theoremstyle{plain}

\theoremstyle{definition}

\theoremstyle{remark}

\usepackage[textsize=tiny]{todonotes}

\begin{document}
	
	\twocolumn[
	\icmltitle{ModulE: Module Embedding for Knowledge Graphs}
	
	
	
	\icmlsetsymbol{equal}{*}
	
	\begin{icmlauthorlist}
		\icmlauthor{Jingxuan Chai$^\textbf{*}$}{}
		\icmlauthor{Guangming Shi}{}
	\end{icmlauthorlist}
	
	
	\icmlcorrespondingauthor{Firstname1 Lastname1}{first1.last1@xxx.edu}
	\icmlcorrespondingauthor{Firstname2 Lastname2}{first2.last2@www.uk}
	
	\icmlkeywords{Machine Learning, ICML}
	
	\vskip 0.3in
	]
	
	
	
	
	\begin{abstract}
		Knowledge graph embedding (KGE) has been shown to be a powerful tool for predicting missing links of a knowledge graph. However, existing methods mainly focus on modeling relation patterns, while simply embed entities to vector spaces, such as real field, complex field and quaternion space. To model the embedding space from a more rigorous and theoretical perspective, we propose a novel general group theory-based embedding framework for rotation-based models, in which both entities and relations are embedded as group elements. Furthermore, in order to explore more available KGE models, we utilize a more generic group structure, module, a generalization notion of vector space. Specifically, under our framework, we introduce a more generic embedding method, \textbf{ModulE}, which projects entities to a module.  Following the method of ModulE, we build three instantiating models: ModulE$_{\mathbb{R},\mathbb{C}}$, ModulE$_{\mathbb{R},\mathbb{H}}$ and ModulE$_{\mathbb{H},\mathbb{H}}$, by adopting different module structures. Experimental results show that ModulE$_{\mathbb{H},\mathbb{H}}$ which embeds entities to a module over non-commutative ring, achieves state-of-the-art performance on multiple benchmark datasets.
		\renewcommand{\thefootnote}{}
		\footnotetext{
			\begin{itemize}
				\item Jingxuan Chai, Guangming Shi are  with School of Artificial Intelligence, Xidian University, Xi'an, Shanxi, 710071, China.
				\item * Corresponding author: Jingxuan Chai.
			\end{itemize}
			}
	\end{abstract}
	
	\section{Introduction}
	Knowledge graphs (KGs) have drawn widespread attraction for their successful use in many downstream tasks, such as question answering \cite{bordes2014open}, semantic parsing \cite{berant2013semantic} and recommendation system \cite{wang2018ripplenet}. Due to the incompleteness problems that a lot of possible triples are missing, link prediction, also known as knowledge graph completion (KGC), has been a fundamental problem. Knowledge graph embedding (KGE), which maps entities and relations from nodes and edges of a graph to low-dimensional representations, has shown promising performance as well as interpretability.

	Recent KGE methods mainly focus on finding relation patterns and modeling each relation as rotation and scaling operation in a vector space, so called rotation-based models. RotatE \cite{sun2019rotate} represents relation as rotation in complex space, constraining the modulus of each rotation embedding to be 1. QuatE \cite{zhang2019quaternion} utilizes hyper-complex representation and models relations as rotations in quaternion space. DensE \cite{lu2020dense} further explores the effect of varying modulus by modeling each relation as rotation and scaling operator on 3D vectors. However, these approaches do not give an insight into the embedding space of entities. To deliver a high-level analysis for overall KGE process, a general KG representation learning framework, which models both relation and entity embedding, is highly demanded.
	
	For the general analysis of relation embedding, NagE \cite{yang2020nage} demonstrates the hidden group structure of relations, and regards them as elements of a group, which is a more theoretical way of modeling hyper-relations of a KG. For entity embedding, existing KGE models generally project entities to some simple vector spaces, without further and systematic discussion.
	
	In fact, a vector space is an Abelian group $V$ over a field $F$, denoted by $V_F$, based on group theory. To construct a general embedding framework, it makes sense if we use the notion of group for embedding space. Furthermore, for a rotation based model that projects entities to vector space $V_F$, the modulus part and orientation part of an embedding vector are the elements of $V$ and $F$ respectively, which suggests that the definitions of vector group and scalar field are essential for generally designing rotation-based embedding models.
	
	To further explore more available models, we should move beyond the notion of vector space. In group theory, a module is the generalization of the notion of vector space, wherein the field of scalars is replaced by a ring. Thus, we can use module as a more generic representation for embedding space. Furthermore, a module is a vector space, if the module is defined on a commutative ring (hence a field). This suggests that a general module-based embedding model can accommodate existing KGE models that project entities to vector space. Moreover, we can utilize the group structures of modules over non-commutative rings to construct a novel embedding model, which has not been studied before.
	
	In this work, we focus on the algebraic structure of embedding vectors. More specifically, we first construct a group embedding framework for rotation-based models, under which we further propose ModulE embedding method based on the notion of module. Finally, we construct 
	three instantiating models by following ModulE: ModulE$_{\mathbb{R},\mathbb{C}}$, ModulE$_{\mathbb{R},\mathbb{H}}$ and ModulE$_{\mathbb{H},\mathbb{H}}$, adopting different module structures. 
	Our contributions are as follows:
	\begin{itemize}
		\item  Given what we know, we are the first to propose a general group theory-based KGE framework that models both entity and relation embeddings as group elements.
		\item Our proposed ModulE embedding method is able to accommodate most of the existing KGE models and instantiate new models by applying different module structures that entities and relations are projected to.
		\item Given what we know, we are the first to use the notion of module for embedding space. Our proposed ModulE$_{\mathbb{H,H}}$ is the first model to map entities to elements of a module over non-commutative ring.
		\item Empirically, our ModulE$_{\mathbb{H,H}}$ model significantly outperforms several state-of-the-art models on three benchmarks. 
	\end{itemize}

	\section{Related Works}
	In this section, we brief several KGE works related to our approach.
	
	\textbf{Distance-based Models} TransE \cite{bordes2013translating} is the opening work for distance-based models, which interprets relation as a translating vector $\bm{r}$ from head entity to tail entity of a triple, i.e. $\bm{h}+\bm{r}=\bm{t}$. Several models are proposed to improve the performance of TransE.
	Specifically, TransH \cite{wang2014knowledge} points that TransE does not perform well on predicting complex relations and proposes a model making entities to have distinct representation given different relations. TransR \cite{lin2015learning} tackles the complex relation problem by  projecting entities and relations to two different spaces. TransD \cite{ji2015knowledge} uses independent projection for each entity and relation, and reduces the amount of parameters compared with TransR.
	
	\textbf{Bilinear Models} These models apply product-based score functions to match latent semantics of entities and relations. Thus, bilinear models are also known as semantic matching models. In RESCAL \cite{2011A}, each relation is embedded as a full rank matrix, where a bilinear score function, $f_r(h,t)=\bm{h}^T\textbf{M}_r\bm{t}$, is adopted. However, the assumption of full rank matrix leads to an overfitting problem. Following approaches use other assumption, such as additional assumption, to address this challenge. DistMult \cite{yang2014embedding} replaces the relational matrix with a diagonal matrix in order to reduce complexity. ANALOGY \cite{liu2017analogical} supposes that $\textbf{M}_r$ is normal. ComplEx \cite{trouillon2017complex} extends DistMult by applying complex space for embedding, which is the first work to introduce complex-valued embedding.
	
	\textbf{Rotation-based Models} RotatE \cite{sun2019rotate} finds that distance-based models are not able to model some relation patterns such as symmetry and proposes a rotation-based model which models relation as rotation in a complex space, i.e. $\bm{t}=\bm{h}\odot\bm{r}$, where $\odot$ denotes the element-wise product between complexes.   
	QuatE \cite{zhang2019quaternion} extends the embedding space of RotatE from complex space to quaternion space.  HAKE \cite{zhang2020learning} decomposes the embedding vector into its modulus part and phase part, and defines two different functions to score the modeling of these two parts. DensE \cite{lu2020dense} represents relation as scaling and rotation in $\mathbb{R}^3$. In general, the relational translation on embedding vectors of rotation-based models consist of two steps: scaling the modulus, followed by rotating the orientation. 
	
	\textbf{Group Embedding} TorusE \cite{ebisu2018toruse} is the first approach to apply the notion of group, which defines embeddings in a compact Lie group, torus. NagE \cite{yang2020nage} finds the hidden group structure of relations in KG and provides a general group embedding recipe, where the relation embeddings are regarded as group elements. Furthermore, the requirement of the non-commutativity of hyper-relations in KG suggests implementing non-Abelian groups for the most general KG tasks. By following its group embedding recipe, NagE proposes two models with non-Abelian groups $SO(3)$ and $SU(2)$. However, there is still a lack of theoretical analysis of entity embedding. Thus, the investigation of group representation theory of entity is highly demanded. 
	
	\section{Background}
	In this section, we show the definitions of some important notions related to vector space under abstract algebra theory.
	\subsection{Vector space}
	\textbf{Definition 1.} Let $V$ be an Abelian group under the operation $+$ and let $F$ be a field. Consider a map, called scalar multiplication:
	\begin{equation}
		F\times V\to V: (\alpha,x) \to \alpha x,
	\end{equation}
	that satisfies properties of compatibility, identity and positive-definiteness (see Appendix \ref{prop_vec} for details).
	
	An Abelian group $V$ for which there is a scalar multiplication map called a vector space over field $F$, denoted by $V_F$. The elements of $F$ are called scalars.
	
	\subsection{Module}
	In a vector space, the set of scalars is a field and acts on the vectors by scalar multiplication, subject to certain axioms. Module is a  generalized notion of vector space with a milder constraint on scalars. 
	
	\textbf{Definition 2.} Let $M$ be be an Abelian group under the operation $+$ and let $R$ be a ring. The map: 
	\begin{equation}
		R\times M\to M: (\alpha,x) \to \alpha x,
	\end{equation}
	is called the scalar multiplication on module, which has the properties as same as the forms of scalar multiplication on vector space (see Appendix \ref{prop_mod} for details). An Abelian group for which there is a scalar multiplication on module is called a  module over $R$. $M$ is a vector space, if $R$ is a field.
	
	\subsection{Field Norm}
	In field theory, a norm is the determinant of a linear transform of a vector space.
	
	\textbf{Definition 3.} Let $K$ be a field and $L$ a finite extension of $K$. The field $L$ is then a finite dimensional vector space over $K$. Multiplication by $\alpha$, an element of $L$:
	\begin{equation}
		L\to L: M_\alpha(x)=\alpha x,
	\end{equation}
	is a $K$-linear transformation of this vector space into itself.
	The norm, $N_{L/K}(\alpha)$, is defined as the determinant of this linear transformation \cite{lidl1983finite}. One simple example is the field norm from complex to real. Complex field $\mathbb{C}$ is a finite extension of real field $\mathbb{R}$. Given $\alpha=a+bi\in\mathbb{C}$, the norm of $\alpha$ is $N_{\mathbb{C}/\mathbb{R}}(a+bi)=a^2+b^2$. 
	
	\subsection{Inner Product Map}
	\textbf{Definition 4.} An inner product on a vector space $V$ over field $F$ is a map:
	\begin{equation}
		V\times V\to F: \langle x,y\rangle\to m,
	\end{equation}
	that satisfies properties of conjugate symmetry, linearity and positive-definiteness (see Appendix \ref{prop_inner} for details).

	\section{Method}
	In this section, we first give some notations related to KGE task and provides a general group theory-based embedding strategy for entity and relation. Then we focus on the rotation-based KGE models, and propose a group embedding framework, under which we further introduce our ModulE embedding method, employing the notion of module. Finally, we construct three example models by following ModulE, with different module structures.  
	
	\subsection{Problem Formulation}
	Let $\mathcal{E}$ denote the set of entities and $\mathcal{R}$ denote the set of relations, then the knowledge graph (KG) $\mathcal{G}$ is the set of factual triples, i.e. $\mathcal{G}=\{(h,r,t)\}$, where $h,t \in \mathcal{E}$ and $r\in \mathcal{R}$. The link prediction task of KG aims to predict missing links between entities based on given facts. To measure the plausibility of candidate triples $(h,r,t)$, a score function is defined as  $f_r(h,t)$. The goal of a KG embedding (KGE) model is to map entities and relations to continuous vector representations. The entity and relation vectors obtained by KGE model make it possible to calculate the score of a triple for KGC.
	
	\subsection{Group Theory based Entity and Relation Embedding}
	
	\textbf{Entity Embedding} Most of the existing KGE models generally project the set of entities into continuous vectors spaces, so that the inner product or distance score functions are available as the score function. 
	In group theory, a vector space is an Abelian vector group over a scalar field. To design KGE models from a high-level perspective, we use the notion of group for entity embedding.
	
	Formally, let $v_e$ denote an entry of an $n$-dimensional entity embedding vector $\bm{v}_e$ for entity $e$, and $v_e$ is an element of a group $G$ which is termed as \textbf{entity group}.
	
	\textbf{Relation Embedding} One of the ways for relation embedding to model relation patterns of a KG is adopting transformation group \cite{yang2020nage}. For relational group embedding, relations are embedded as group elements, which are parameterized by certain group parameters. Then each relation acts as a mapping from one entity representation to another:
	\begin{equation}
		G \to G: T_{v_r}(v_e)=v_e', 
	\end{equation}
	where $T_{v_r}(\cdot)$ is the element of a transformation group $T$ parameterized by $v_r$ of relation $r$, and $v_e'$ denotes an entry of transformed entity embedding $\bm{v}_e'$. For the multi-dimensional mapping for relation embedding, we have: 
	\begin{equation}
		G^n\to G^n: T_{\bm{v}_r}(\bm{v}_e)=\bm{v}_e',
	\end{equation} 
	where $T_{\bm{v}_r}\in T^n$ denotes an $n$-tuple of elements of $T$ parameterized by $\bm{v}_r$. We call $T$ \textbf{relation group}.
	
	\subsection{A Group Embedding Framework for Rotation-Based Models}
	
	\begin{table*}[t]
		\caption{Examples of group embedding for rotation based model. NagE* represents the SO3E \cite{yang2020nage} model.}
		\label{eg-table}
		\vskip 0.15in
		\begin{center}
			\renewcommand{\arraystretch}{1.2}
			\arrayrulewidth0.7pt
			\begin{tabular}{|c|c|c|c|c|c|c|}
				\hline
				\textbf{Model} & \textbf{Scalar} $S$ & \textbf{Vector} $V$ & \textbf{Scaling} $T_S$ & \textbf{Rotation} $T_V$ & \textbf{Norm} &\textbf{Score}\\
				\hline
				DistMult \cite{yang2014embedding} & $\mathbb{R}$ & $\mathbb{R}$ & $GL(1)$ & $V^f$ &$N_{\mathbb{R}/\mathbb{R}}(\cdot)$& distance \\\hline
				RotatE \cite{sun2019rotate}& $\mathbb{R}$ & $\mathbb{C}$ & $S^f$ & $U(1)$ &$N_{\mathbb{C}/\mathbb{R}}(\cdot)$&distance \\\hline
				HAKE \cite{zhang2020learning} & $\mathbb{R}$ & $\mathbb{R}^2$ & $GL(1)$ & $SO(2)$ &$N_{\mathbb{R}^2/\mathbb{R}}(\cdot)$& distance \\\hline
				NagE* \cite{yang2020nage} & $\mathbb{R}$ & $\mathbb{R}^3$ & $S^f$ & $SO(3)$ &$N_{\mathbb{R}^3/\mathbb{R}}(\cdot)$& distance \\\hline
				QuatE \cite{zhang2019quaternion}& $\mathbb{R}$ & $\mathbb{H}$ & $S^f$ & $U_\mathbb{H}(1)$ &$N_{\mathbb{H}/\mathbb{R}}(\cdot)$& cosine \\\hline
				\textbf{ModulE}$_{\mathbb{R},\mathbb{C}}$& $\mathbb{R}$ & $\mathbb{C}$ & $GL(1)$ & $U(1)$ &$N_{\mathbb{C}/\mathbb{R}}(\cdot)$& cosine \\\hline
				\textbf{ModulE}$_{\mathbb{R},\mathbb{H}}$& $\mathbb{R}$ & $\mathbb{H}$ & $GL(1)$ & $U_\mathbb{H}(1)$ &$N_{\mathbb{H}/\mathbb{R}}(\cdot)$& cosine\\\hline
				\textbf{ModulE}$_{\mathbb{H},\mathbb{H}}$& $\mathbb{H}$ & $\mathbb{H}$ & $U_\mathbb{H}(1)$ & $U_\mathbb{H}(1)$ &$N_{\mathbb{H}/\mathbb{R}}(\cdot)$& cosine\\\hline
			\end{tabular}
		\end{center}
		\vskip -0.1in
	\end{table*}
	
	Rotation-based models represent relations as rotations in embedding spaces, such as RotatE \cite{sun2019rotate}, QuatE \cite{zhang2019quaternion} and DensE \cite{lu2020dense}, which are capable of modeling multiple relation patterns. The relation embedding of a rotation based model generally act as two operations on an entity vector: rotating its orientation and scaling its length. 
	
	For rotation-based models, general descriptions of the modulus part and orientation part of embedding vector are required. In this section, we propose a group embedding framework to represent the relational variation of vector and scalar on an embedding space, by using the aforementioned group embedding strategy.
	
	\textbf{Entity Embedding} An entity $e$ is mapped to a scalar embedding $\bm{s}_e$ and a vector embedding $\bm{v}_e$. Let $s_e,v_e$ denote the entries of $\bm{s}_e,\bm{v}_e$. A \textbf{scalar group} $S$ and \textbf{vector group} $V$ are defined, for which we have $s_e\in S, v_e\in V$. The modulus of each $v_e$ is constrained to be $1$. For most of the existing KGE models, $S$ is real field, and $V$ is a vector space, such as $\mathbb{C}$ or $\mathbb{H}$.
	
	\textbf{Relation Embedding} An relation $r$ is mapped to a scaling parameter $\bm{s}_r$ and a rotation parameter $\bm{v}_r$. Let $s_r, v_r$ denote the entries of $\bm{s}_r,\bm{v}_r$. A \textbf{scaling group} $T_S$ and \textbf{rotation group} $T_V$ are defined, for which we have $T_{s_r}\in T_S, T_{v_r}\in T_V$, where $T_{s_r}$ and $T_{v_r}$ are the group elements parameterized by $s_r$ and $v_r$. The multi-dimensional representations are $T_{\bm{s}_r}$ and $T_{\bm{v}_r}$. Thus, the transformed scalar and vector can be written as $\bm{s}_e'=T_{\bm{s}_r}(\bm{s}_e)$ and $\bm{v}_e'=T_{\bm{v}_r}(\bm{v}_e)$:
	
	\textbf{Combination of Modulus and Rotation} The scalar embedding and vector embedding are combined by applying scalar multiplication map:
	\begin{equation}
		S\times V\to V: (s_e,v_e)\to s_e\cdot v_e.
	\end{equation}
	The scalar multiplication between  $\bm{s}_e$ and $\bm{v}_e$ is denoted as $\bm{s}_e \odot \bm{v}_e$, where $\odot$ is the element-wise scalar multiplication. 
	
	\textbf{Score function} For a triple $(h,r,t)$, let $\bm{h}=\bm{s}_h \odot \bm{v}_h$ and $\bm{t}=\bm{s}_t \odot \bm{v}_t$ denote the combined embeddings of head and tail entity. Then we obtain the transformed head embedding by combining its transformed scalar and vector part:
	\begin{equation}
		\bm{h}'=T_{\bm{s}_r}(\bm{s}_h)\odot\ T_{\bm{v}_r}(\bm{v}_h)
	\end{equation}
	The score function is set up to measure the similarity between transformed head entity and object tail entity in the form of a degenerate map from the Cartesian product of $G^n$ and $G^n$ to real field:
	\begin{equation}
		G^n\times G^n\to \mathbb{R}: f(\bm{h}',\bm{t})=s.
	\end{equation}
	If $V$ is a vector space over $\mathbb{R}$, one could use degenerate maps on $V$ as score function, such as norm map and distance function.
	
	\textbf{Examples of Group Embedding for Rotation-Based Models}  
	We demonstrate some embedding cases instantiated by our group embedding framework,  where we use $\mathbb{R}$ for scalar groups, and several vector spaces for vector groups, since the field norm and inner product are already defined on these vector spaces over $\mathbb{R}$.
	
	One could use complex field $\mathbb{C}$ for vector group $V$, linear map group $GL(1)$ for scaling group $T_S$, and $SO(2)$ group for rotation group $T_V$. This embedding model corresponds to RotatE \cite{sun2019rotate}. One could use quaternion space $\mathbb{H}$ for vector group $V$, fixed-point map group\footnote{The fixed-point subgroup $S^f$ of an automorphism $f$ of a group $S$ is the subgroup of $S$: $S^f=\{s\in S|f(s)=s\}.$} $S^f$ for scaling group $T_S$, and quaternion rotation group\footnote{The quaternion rotation group is the group: $U_\mathbb{H}(1)=\{ Q\in GL(1,\mathbb{H}), \overline{Q} Q=1 \}$.} $U_\mathbb{H}(1)$ for rotation group $T_V$. This embedding model corresponds to QuatE \cite{zhang2019quaternion}. Example models instantiated by our group embedding framework that correspond to 5 studied models are shown in Table \ref{eg-table} in detail.
	
	\subsection{Module Embedding for Rotation Based Model}
	As Table \ref{eg-table} shown, traditional KGE models project entities to vector spaces. To explore more available models, we move beyond the vector space for entity embedding and adopt the notion of module. For our embedding framework in Section 4.3, we extend the embedding space from vector spaces to modules, and we get \textbf{ModulE: the module embedding method}, a more general embedding framework.  However, the variation of the commutativity of the scalar ring a module defined on makes significant effect on two parts of our general group embedding model, which are discussed below.
	
	\textbf{Norm Map of Module} A Field norm obtains a normalized representation of a vector. One could use the norm to calculate the length (modulus) of elements of vector space. For example, we have $N_{\mathbb{H}/\mathbb{R}}(a+bi+cj+dk)=a^2+b^2+c^2+d^2$, where $\mathbb{H}$ is quaternion space.
	
	As for norm map of $n$-tuple on a vector space $V$, existing KGE models generally use the $L_p$ norm function mapping the $n$-dimensional embedding vectors to real numbers:
	\begin{equation}
		V^n\to \mathbb{R}: L_p(\bm{x})=\left(\sum_{i\in n} N_{V/\mathbb{R}}^p(x_i) \right)^{1/p},
	\end{equation}
	for score function.
	
	In fact, field norm is the determinant of a linear transformation of a vector space (see \textbf{Definition 3}.), which means there is no norm defined on module over non-commutative ring. To maintain the generality of ModulE, a general norm map from a module to a real number is required:
	\begin{equation}
		M\to \mathbb{R}: g(x)=m.
	\end{equation}
	Note that if module $M$ is a vector space over $\mathbb{R}$, one could use $N_{M/\mathbb{R}}(\cdot)$ for general norm map. 
	Then the $n$-dimensional mapping for module embedding can be written as:
	\begin{equation}
		M^n\to \mathbb{R}: G_p(\bm{x})=\left(\sum_{i\in n} N_{M/\mathbb{R}}^p(x_i) \right)^{1/p}.
	\end{equation}
	We call $G_p(\cdot)$ the general $n$-dimensional norm map on module $M$, which is equivalent to $L_p$ norm if $M$ is a vector space over $\mathbb{R}$.
	
	\textbf{Inner Product Map of Module} Some of the existing models use inner product to calculate the similarity between entity embeddings. Inner product is a map from vector space to a field, hence no ``inner product like'' operation defined on module over non-commutative ring. To obtain the similarity  between elements of modules in the form of real number, we need to define a general inner product map from the Cartesian product of $M$ and $M$ to real field:
	\begin{equation}
		M\times M \to \mathbb{R}: p(x,y)=m.
	\end{equation}
	Note that if module $M$ is a vector space over $\mathbb{R}$, one could use inner product map $\langle \cdot,\cdot\rangle$ as the degenerate map.
	
	\subsection{Embedding Cases of ModulE}
	Following the embedding method of ModulE, we demonstrate three instantiating models using different module structures. 
	
	\textbf{Module Embedding over Field} 
	We first construct two models with modules over real field $\mathbb{R}$ (hence vector spaces over real field). Specifically,
	we propose \textbf{ModulE$_{\mathbb{R},\mathbb{C}}$} where we use complex field $\mathbb{C}$ for vector group, $GL(1)$ for scaling group and $U(1)$ for rotation group, and \textbf{ModulE$_{\mathbb{R},\mathbb{H}}$} where we use quaternion space $\mathbb{H}$ for vector group, $GL(1)$ for scaling group and $U_\mathbb{H}(1)$ for rotation group.
	
	\textbf{Module Embedding over Non-Commutative Ring} We also construct a model with module over a non-commutative ring, the quaternion ring. For our third proposed model \textbf{ModulE$_{\mathbb{H},\mathbb{H}}$}, we use quaternion ring $\mathbb{H}$ for both scalar group and vector group, the quaternion rotation group $U_\mathbb{H}(1)$ for both scaling group and rotation group. 
	
	Note that the normal scalar multiplication ``$\cdot$'' is not available as a combination operation between hyper-complex scalar and vector. Thus, we adopt the quaternion multiplication ``$\times$''. Then the combination of scalar part $\bm{s}_e$ and vector part $\bm{v}_e$ of an entity can be written as $\bm{s}_e\otimes\bm{v}_e$, where $\otimes$ denotes the element-wise quaternion multiplication.
	
	
	\subsection{Score Function and Train Loss}
	For each triple $(h,r,t)$, we define the score function using inner product (cosine similarity) between transformed head entity and tail entity vector:
	\begin{equation}
		f_r(h,t)=\langle\bm{h}',\bm{t}\rangle.
	\end{equation}
	The inner product maps and norm maps are already defined on the vector spaces of ModulE$_{\mathbb{R},\mathbb{C}}$ and ModulE$_{\mathbb{R},\mathbb{H}}$, but not on the module of ModulE$_{\mathbb{H},\mathbb{H}}$. Thus we use the inner product of quaternion space for general inner product map and $N_\mathbb{H,H}$ for general norm map of ModulE$_{\mathbb{H},\mathbb{H}}$. 
	
	We regard the KGC task as a classification problem by following previous work \cite{trouillon2017complex}, and apply regularized logistic loss to train our models:
	\begin{equation}
		\mathcal{L}=\sum_{(h,r,t)\in\mathcal{G}}\left(\sum_{t'\in\mathcal{E}}{\rm log}(1+{\rm exp}(-y_tf_r(h,t'))\right) + \Phi,\\
	\end{equation}
	where $y_t$ is the binary indicator related to $t$ and $t'$. Specifically, $y_t=1$ if $t=t'$, otherwise $y_t=0$. Here we use general $n$-dimensional norm map $G_p(\cdot)$ on module $M$ with regularization rates $\lambda_1, \lambda_2, \lambda_3$ to combat overfitting: 
	\begin{equation}
		\Phi=\sum_{(h,r,t)\in\mathcal{G}}\lambda\left(\lambda_1G_p(\bm{h})+\lambda_2G_p(\bm{r})+\lambda_3G_p(\bm{t})\right), 
	\end{equation}
	where $\bm{r}\in M^n$ is the $n$-tuple module representation of relation $r$, $\lambda$ is the  regularization multiplier.
	
	\section{Experiments}
	To validate the effectiveness of our proposed models, we conduct experiments on several widely used datasets, including FB15k-237 \cite{dettmers2018convolutional}, WN18RR \cite{toutanova2015observed} and YAGO3-10 \cite{mahdisoltani2014yago3}.
	
	\subsection{Datasets}
	FB15k-237 and WN18RR are subsets of FB15k \cite{bordes2013translating} and WN18 \cite{bordes2013translating} respectively. Both FB15k and WN18 suffer from test leakage problem \cite{toutanova2015observed}, on which KGE models typically perform well on. Therefore, we do not use them in our experiments for one can attain state-of-the-art performance even using a simple rule based model. To make the task of link prediction more challenging, we use FB15k-237 and WN18RR where the inverse relations are deleted from their original test sets of FB15k and WN18. We also use the YAGO3-10 dataset, a subset of YAGO3 \cite{mahdisoltani2014yago3}. YAGO3-10 consists of a large collection of triplets from multilingual Wikipedia.
	
	\subsection{Evaluation Protocols}
	With respect to the benchmark dataset, the objective metric for comparative analyses are: mean reciprocal rank (MRR) and Hits@K (Hits@1, Hits@3 and Hits@10). Mean reciprocal rank is the average triples over all candidate entities. Hits@K evaluates the percentage of times a true triple is ranked at top K of predicted results. Here we apply the  \textsc{BOTTOM} \cite{sun2020A} evaluation protocol, where the correct triple is placed at the end of a list of triples with same scores. This is a more strict setting compared with the filtered setting \cite{bordes2013translating}, where all given true triples are removed from candidate set except for the current test triple.

	\subsection{Implementation Details for ModulE}
	We implement our models in Pytorch and tested it on a single GPU. We use the Adagrad \cite{duchi2011adaptive} optimizer for learning with learning rate of 0.1. For experiments on WN18RR, we further apply an exponentially decaying learning schedule with a decay rate of 0.1. 
	
	To obtain best models, we select hyperparameters by early stopping on the validation sets. In general, the embedding dimension multiplier $k$ is selected from \{32, 64, 128, 256, 512\} (The embedding multiplier is the variable multiplier of embedding size. Specifically, the embedding size for ModulE$_\mathbb{H,H}$ is $7k$). The regularization multiplier $\lambda$ is selected from \{5e-3, 1e-2, 3e-2, 4e-2, 4.5e-2, 5e-2, 6e-2, 7e-2, 8e-2\}. Regularization rate $\lambda_1, \lambda_2, \lambda_3$ are searched from \{0.5, 1.0, 1.5, 2.0, 2.5, 3.0\}. For the $p$-parameter of the general norm map $G_p(\cdot)$, we test our models with $p=2$ and $p=3$ respectively. 
	
	The best hyperparameter settings for each benchmark dataset are shown in Appendix \ref{hyper_para}.
	
	\subsection{Baselines}
	We compare the performance of our proposed ModulE models with multiple state-of-the art KGE models, including TransE \cite{bordes2013translating}, DistMult \cite{yang2014embedding}, ComplEx \cite{trouillon2017complex}, RotatE \cite{sun2019rotate}, QuatE \cite{zhang2019quaternion}, NagE \cite{yang2020nage},  DensE \cite{lu2020dense}, HopfE \cite{bastos2021hopfe},
	GprQ8 \cite{yang2021knowledge}.
	
	\begin{table*}[t]
		\caption{Link Prediction results on FB15K-237, WN18RR and YAGO3-10. Best results are in bold and second best results are underlined. NagE* represents the SO3E \cite{yang2020nage} model. QuatE$\dagger$ represents the results are taken from \cite{bastos2021hopfe}.}
		\label{MRR-table}
		\vskip 0.15in
		\begin{center}
			\begin{tabular}{ccccccccccccc}
				\toprule
				&\multicolumn{4}{c}{\textbf{FB15K-237}}&\multicolumn{4}{c}{\textbf{WN18RR}}&\multicolumn{4}{c}{\textbf{YAGO3-10}}\\
				Model & MRR & H@1 & H@3 & H@10 & MRR & H@1 & H@3 & H@10 & MRR& H@1 & H@3 & H@10\\
				\midrule
				TransE & 0.294 & - & - & 0.465 & 0.226 & - & - & 0.501 & - & - & - & -\\
				DistMult& 0.241 & 0.155 & 0.263 & 0.419 & 0.430 & 0.390 & 0.440 & 0.490 & 0.340 & 0.240 & 0.380 & 0.540\\
				ComplEx & 0.247 & 0.158 & 0.275 & 0.428 & 0.440 & 0.410 & 0.460 & 0.510 & 0.360 & 0.260 & 0.400 & 0.550\\
				RotatE & 0.338 & 0.241 & 0.375 & 0.533 & 0.476 & 0.428 & 0.492 & 0.571 & - & - & - & -\\
				QuatE$\dagger$ & 0.311 & 0.221 & 0.342 & 0.495 & 0.481 & 0.436 & \underline{0.500} & 0.564 & - & - & - & -\\
				NagE* & 0.340 & 0.244 & 0.378 & 0.530 & 0.477 & 0.432 & 0.493 & \underline{0.574} & - & - & - & -\\
				DensE & 0.349 & 0.256 & - & - & \underline{0.491} & \underline{0.443} & - & - & 0.541 & 0.465 & - & -\\
				HopfE & 0.343 & 0.247 & 0.379 & 0.534 & 0.472 & 0.413 & \underline{0.500} & \textbf{0.586} & 0.529 & 0.438 & 0.586 & 0.695\\
				GrpQ8 & \underline{0.355} & \underline{0.262} & - & - & 0.474 & 0.435 & - & - &  - & - & - & -\\\midrule
				ModulE$_{\mathbb{R},\mathbb{C}}$ & 0.343 & 0.260 & 0.384 & 0.529 & 0.467 & 0.434 & 0.478 & 0.525 & 0.551 & 0.471 & 0.600 & 0.701\\
				ModulE$_{\mathbb{R},\mathbb{H}}$ & 0.351 & 0.259 & \underline{0.385} & \underline{0.539} & 0.478 & 0.437 & 0.489 & 0.555 & \underline{0.564} & \underline{0.485} & \underline{0.611} & \underline{0.707}\\
				ModulE$_{\mathbb{H},\mathbb{H}}$ & \textbf{0.361} & \textbf{0.267} & \textbf{0.398} & \textbf{0.555} & \textbf{0.492} & \textbf{0.451} & \textbf{0.506} & 0.568 & \textbf{0.578} & \textbf{0.502} & \textbf{0.620} & \textbf{0.713}\\
				\bottomrule
			\end{tabular}
		\end{center}
		\vskip -0.1in
	\end{table*}
	
	\section{Results and Analysis}
	\subsection{Performance of Link Prediction}
	The empirical results on FB15k-237, WN18RR and YAGO3-10 are reported in Table \ref{MRR-table}. Our proposed ModulE$_{\mathbb{H,H}}$ significantly outperforms previous state-of-the-art models on all metrics on benchmark datasets, apart from the Hits@10 metric on WN18RR. 
	
	FB15k-237 has the most relations and least entities, compared with WN18RR and YAGO3-10, hence has the most complex relation types. Our ModulE$_{\mathbb{H,H}}$ model achieves the largest margins against most of the previous methods on FB15k-237, compared with other benchmarks. This suggests that hyper-complex scaling and rotation are useful in  modeling complex relation patterns.
	
	WN18RR dataset contains a large proportion of symmetric relations, such as \textit{similar\underline{~~}to, also\underline{~~}see}, and hierarchical relations, such as \textit{hypernym, has\underline{~~}part}. ModulE$_{\mathbb{H,H}}$ scores the best on MRR, Hits@1 and Hits@3 on WN18RR, indicating our model is capable of modeling these two relation types.
	
	YAGO3-10 dataset has entities with high relation-specific indegree \cite{2017Convolutional}, which makes it difficult to predict missing entities for a triple in YAGO3-10. For example, the missing triple \textit{(?, hasGender, male)} has over 1,000 answers, making the link prediction task more challenging. All of our three ModulE models outperform the current state-of-the-art models. Specifically, ModulE$_{\mathbb{H,H}}$ surpasses HopfE \cite{bastos2021hopfe} by $9.26\%, 14.61\%, 5.80\%, 2.52\%$ on MRR, Hits@1, Hits@3, Hits@10 respectively.
	
	Moreover, we compare the performance of our models with two existing models which apply the notion of group. For the most representative group theory-based KGE model, NagE (SO3E \cite{yang2020nage}), our ModulE$_{\mathbb{H,H}}$
	which utilizes not only non-Abelian group for relation embedding, but also module over non-commutative ring for entity embedding, outperforms NagE on almost all the metrics by a huge margin. Notably, ModulE$_{\mathbb{H,H}}$ achieves $6.2\%$ higher MRR on FB15k-237 than NagE. Even our prototype ModulE models, ModulE$_{\mathbb{R,C}}$ and ModulE$_{\mathbb{R,H}}$, show better performance. We also demonstrate the results of GrapQ8 \cite{yang2021knowledge} which applies the notion of groupoid for embedding. ModulE$_{\mathbb{H,H}}$ consistently beats this model, and achieves $4.23\%$ higher MRR on WN18RR.
	
	\begin{table}[t]
		\small
		\caption{Results of ablation studies on FB15K-237 and WN18RR. ``Scalar'' represents the ModulE models with varying modulus only. ``Vector'' represents the ModulE models with varying orientation only. ``Both'' represents the original ModulE models.}
		\label{ablation-table}
		\vskip 0.15in
		\renewcommand{\arraystretch}{1.2}
		\begin{center}
			\begin{tabular}{|c|c|c|c|c|c|}
				\hline
				\multicolumn{2}{|c|}{\multirow{2}{*}{ModulE models}}&\multicolumn{2}{c|}{\textbf{FB15k-237}}&\multicolumn{2}{c|}{\textbf{WN18RR}}\\
				\cline{3-6}
				\multicolumn{2}{|c|}{}& MRR & H@1 &MRR &H@1\\\hline
				\multirow{3}{*}{ModulE$_{\mathbb{R},\mathbb{C}}$} 
				&Scalar & 0.331 & 0.245 & 0.459 & 0.427 \\\cline{2-6}
				&Vector & 0.339 & 0.253 & 0.463 & 0.431\\\cline{2-6}
				&Both & \textbf{0.343} & \textbf{0.260} & \textbf{0.467} & \textbf{0.434} \\\hline
				\multirow{3}{*}{ModulE$_{\mathbb{R},\mathbb{H}}$} 
				&Scalar & 0.331 & 0.245 & 0.459 & 0.427\\\cline{2-6}
				&Vector & 0.347 & \textbf{0.259} & 0.463 & 0.437 \\\cline{2-6}
				&Both & \textbf{0.351} & \textbf{0.259} & \textbf{0.478} & \textbf{0.441} \\\hline
				\multirow{3}{*}{ModulE$_{\mathbb{H},\mathbb{H}}$} 
				&Scalar & 0.352 & 0.262 & 0.469 & 0.440 \\\cline{2-6}
				&Vector & 0.347 & 0.259 & 0.463 & 0.437 \\\cline{2-6}
				&Both & \textbf{0.361} & \textbf{0.267} & \textbf{0.492} & \textbf{0.451} \\\hline
			\end{tabular}
		\end{center}
		\vskip -0.1in
	\end{table}
	
	\begin{table}[t]
		\small
		\caption{MRR results of each relation type for RotatE \cite{sun2019rotate} and ModulE$_\mathbb{H,H}$ on WN18RR. Best results are in bold.}
		\label{relation-table}
		\vskip 0.15in
		\begin{center}
			\begin{tabular}{lcl}
				\toprule
				\textbf{Relation Name} & \textbf{RotatE} &  \textbf{ModulE}$_\mathbb{H,H}$\\
				\midrule
				hypernym  & 0.148 & \textbf{0.179} ($+17.3\%$)\\
				derivationally\underline{~~}related\underline{~~}form & 0.947 & \textbf{0.962} ($+3.3\%$)\\
				member\underline{~~}meronym & \textbf{0.232} & \textbf{0.232}  ($+0\%$)\\
				has\underline{~~}part & 0.184 & \textbf{0.216}  ($+17.4\%$)\\
				instance\underline{~~}hypernym & 0.318 & \textbf{0.418}  ($+31.4\%$)\\
				synset\underline{~~}domain\underline{~~}topic of & 0.340 & \textbf{0.380}  ($+11.8\%$)\\
				also\underline{~~}see& 0.585 & \textbf{0.643}  ($+9.9\%$)\\
				verb\underline{~~}group & 0.943 & \textbf{0.968}  ($+2.7\%$)\\			 
				member\underline{~~}of\underline{~~}domain\underline{~~}region & 0.200 & \textbf{0.329}  ($+64.5\%$)\\
				member\underline{~~}of\underline{~~}domain\underline{~~}usage & 0.318 & \textbf{0.324}  ($+1.9\%$)\\
				similar\underline{~~}to & \textbf{1.000} & \textbf{1.000}  ($+0\%$)\\
				\bottomrule
			\end{tabular}
		\end{center}
		\vskip -0.1in
	\end{table}
	
	\subsection{Ablation Studies}
	In order to examine the effectiveness on each part of our models, we conduct ablation studies on the scalar parts and vector parts of  ModulE$_{\mathbb{R,C}}$, ModulE$_\mathbb{R,H}$ and ModulE$_{\mathbb{H,H}}$. 
	Table \ref{ablation-table} shows the MRR and Hits@1 results on the test sets of FB15K-237 and WN18RR.
	
	We observe that the combination of scalar and vector parts of ModulE models do improve the performance compared to the versions using only one of these parts. For ModulE$_{\mathbb{H,H}}$, the combination model improves the MRR performance by $4.03\%$ on FB15k-237 , and $6.26\%$ on WN18RR,  compared with its vector version.
	
	To evaluate the effectiveness of hyper-complex number against real number for scalar, we compare the performance of ModulE$_\mathbb{R,H}$ and ModulE$_\mathbb{H,H}$.  The MRR performance are lifted by $2.85\%$ and $2.92\%$ on FB15k-237 and WN18RR, which suggests that module over non-commutative ring is more capable of modeling the embedding space than vector space. 
	
	We also see that the vector models outperform the scalar models for ModulE$_{\mathbb{R,C}}$ and ModulE$_{\mathbb{R,H}}$. However, things are opposite for ModulE$_{\mathbb{H,H}}$. In fact, ModulE$_{\mathbb{R,C}}$ and ModulE$_{\mathbb{R,H}}$ both degrade to DistMult \cite{yang2014embedding} model, if we only use the scalar parts of them. It is obvious that real field is lack of embedding capability, compared with complex field and quaternion space. Meanwhile, the scalar version consists of both scaling and rotation transformation, thus performs better than the vector model. 
	
	\subsection{Performance per Relation} 
	To give an insight into the performance gain, we report the performance per relation. Table \ref{relation-table} summarizes the MRR for each relation of ModulE$_{\mathbb{H,H}}$ on the test set of WN18RR, compared with RotatE \cite{sun2019rotate}. We can see that our model gains significant improvement of MRR on most of the relations. Especially for \textit{member\underline{~~}of\underline{~~}domain\underline{~~}region} and \textit{instance\underline{~~}hypernym}, the performance are improved by $64.5\%$ and $31.4\%$. As shown in Table \ref{relation-table}, the improvement of some of the relations are achieved by our model while not sacrificing the performance of other relations.

	\subsection{Embedding Efficiency}
	\begin{figure}[t]
		\begin{center}
			\centerline{\includegraphics[width=\columnwidth]{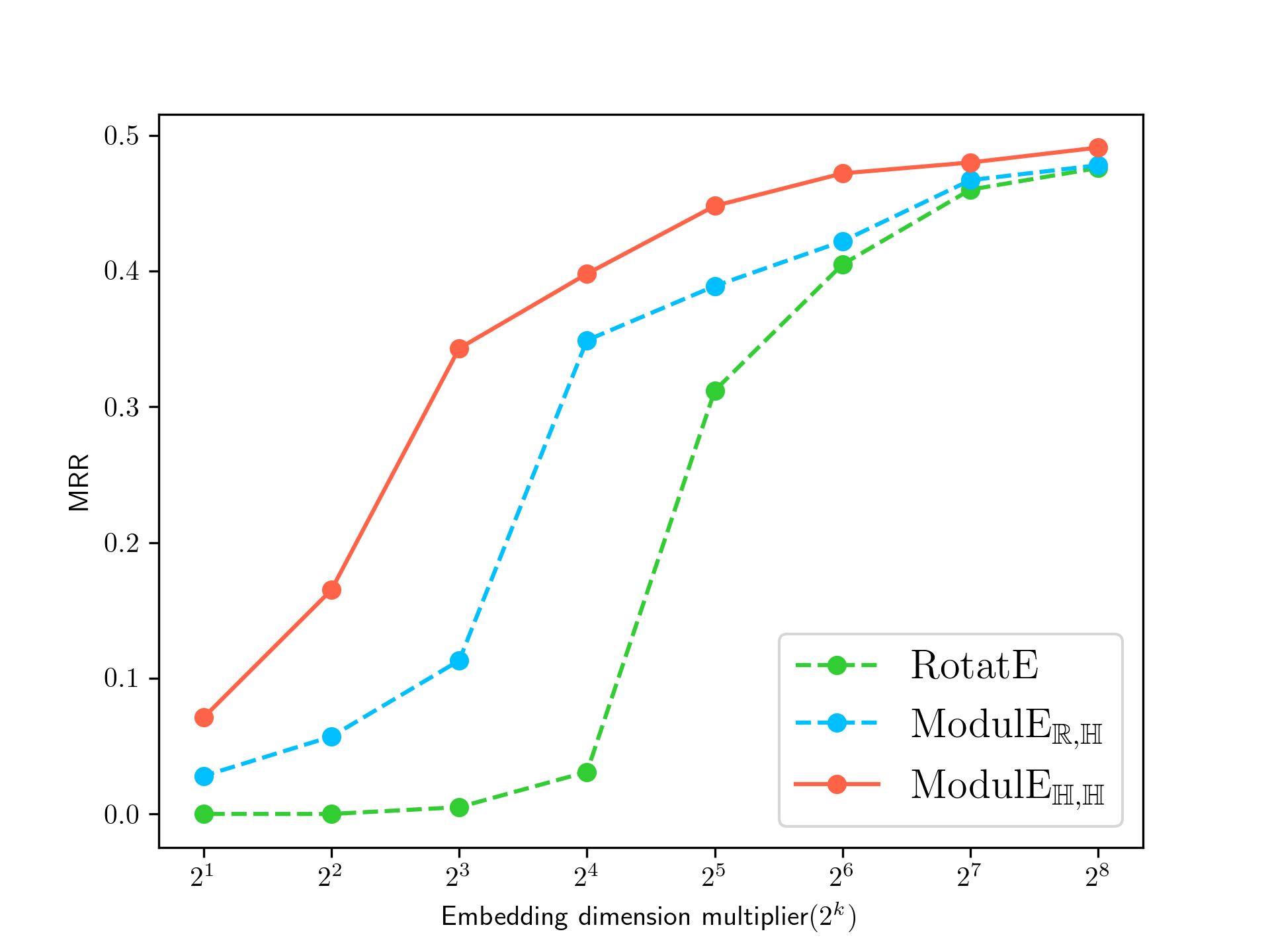}}
			\caption{MRR results of varying embedding size for RotatE \cite{sun2019rotate}, ModulE$_{\mathbb{R,H}}$ and ModulE$_{\mathbb{H,H}}$ on WN18RR.}
			\label{emb_eff_wn}
		\end{center}
		\vskip -0.2in
	\end{figure}
	We further explore the effect of varying embedding size on performance. The MRR results on WN18RR test set of two ModulE models and one studied model, RotatE \cite{sun2019rotate}, are shown in Figure \ref{emb_eff_wn}. With decrease in embedding size, ModulE$_{\mathbb{H,H}}$ perform a much lower declining rate of MRR. Notably,  ModulE$_{\mathbb{H,H}}$ significantly outperform  RotatE and ModulE$_{\mathbb{R,H}}$ under low-dimensional settings. This suggests that models adopting module with hyper-complex scalar is more capable of modeling knowledge graph and is effective even with small embedding size.

	\subsection{Rate of Convergence}
\begin{figure}[t]
		\begin{center}
			\centerline{\includegraphics[width=\columnwidth]{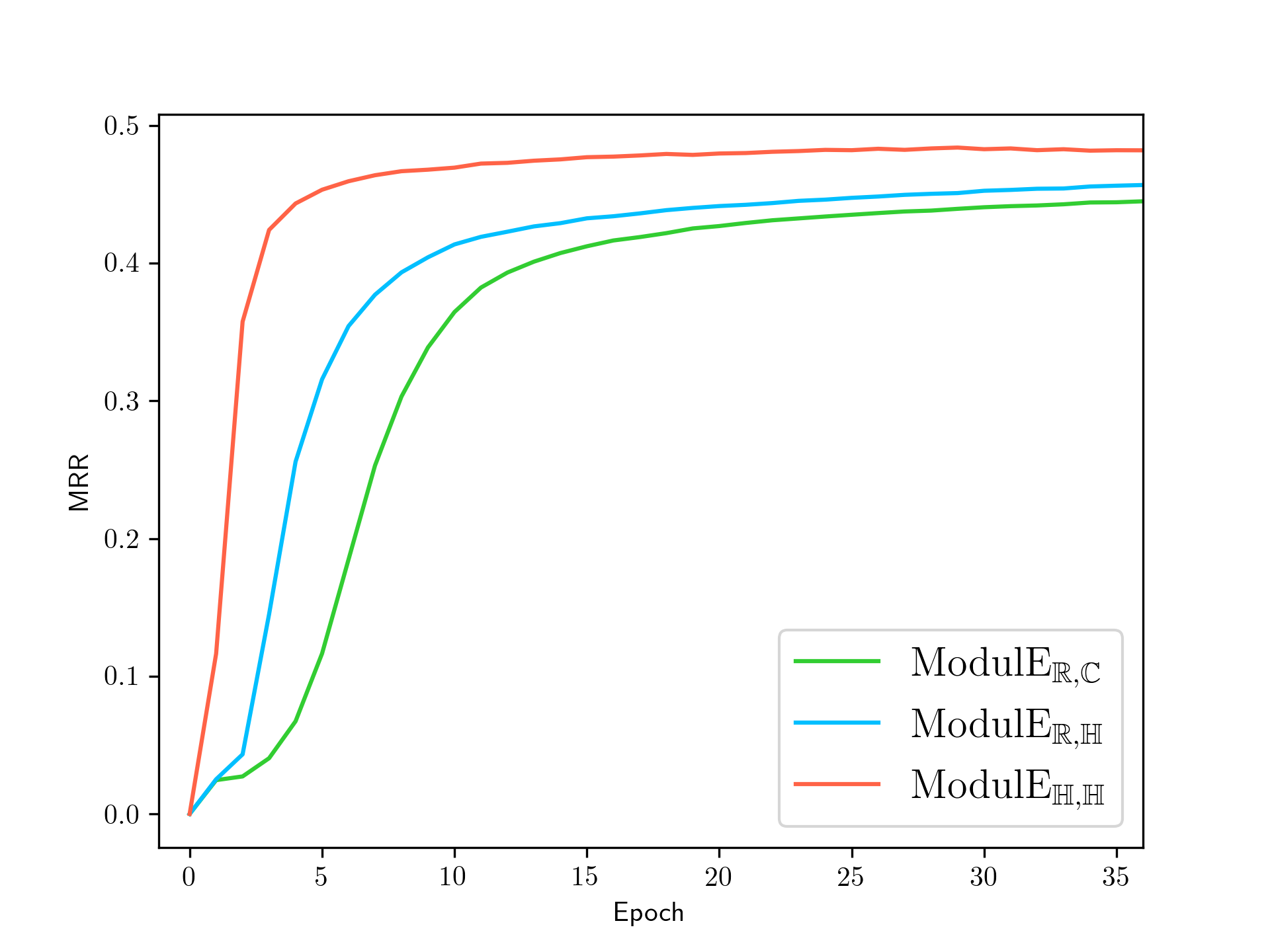}}
			\caption{MRR convergence rate per epoch of three proposed ModulE models on WN18RR.}
			\label{exp_wn}
		\end{center}
		\vskip -0.2in
	\end{figure}

	Figure \ref{exp_wn} shows the MRR results on WN18RR test set per epoch achieved by our three ModulE models. We demonstrate the performance variation of the starting 30 epochs.
	
	We can see that all of them show outstanding convergence rates. Notably, ModulE$_{\mathbb{H,H}}$ shows a fast convergence rate and converge to best MRR performance in 23 epoch, compared with ModulE$_{\mathbb{R,C}}$ and ModulE$_{\mathbb{R,H}}$ (In fact, they both reach best MRR in around 500 epoch on average).
	
	This provides evidence that our proposed model, which utilizes module over non-commutative ring for scalar group, can facilitate the interactions between parameters.  
	
	\section{Conclusion and Future Work}
	In this paper, we propose a novel, theoretically based, general group theory-based embedding method, ModulE, adopting the notion of module. ModulE is capable of accommodating almost all rotation-based KGE models. Following the method of ModulE, we further propose three instantiating models with different module structures. Our ModulE$_{\mathbb{H,H}}$ model, which uses a module over a non-commutative ring, achieves state-of-the-art performance for link prediction task on multiple benchmark datasets. Extensive experiments demonstrate the embedding effectiveness as well as fast convergence rate of ModluE$_{\mathbb{H,H}}$.
	
	Our work provides a general embedding strategy from a theoretical perspective and utilizes a more general group structure for embedding. However, the intrinsic connection between the entity set of a KG and module structure is still in demand. It is required to find the fixed attribute of each entity and relate them to certain group structure.
	
	For future work, we prepare to look into more algebraic knowledge related to group theory, such as field theory, category theory, with the purpose of giving a more general and intuitive framework that integrates both entity and relation of a triple of KG, even for a general triple.
	
	\bibliography{example_paper}
	\bibliographystyle{icml2022}

	\newpage
	\appendix
	\onecolumn
	\section{Full definition of vector space}
	\label{prop_vec}
	Let $V$ be an Abelian group under the operation $+$ and let $F$ be a field. Consider a map, called scalar multiplication:
	\begin{equation}
		F\times V\to V:\quad (\alpha,x) \to \alpha x,
	\end{equation}
	such that for all $x,y\in V$ and all $\alpha,\beta\in F$ it has the properties:
	\begin{itemize}
		\item Distributivity of scalar multiplication with respect to vector addition:
		\begin{equation}
			\alpha(x+y)=\alpha x+\alpha y.
		\end{equation}
		\item Distributivity of scalar multiplication with respect to field addition:
		\begin{equation}
			(\alpha+\beta)x=\alpha x + \beta x.
		\end{equation}
		\item Compatibility of scalar multiplication with field multiplication:
		\begin{equation}
			\alpha(\beta x)=(\alpha\beta)x.
		\end{equation}
		\item Identity element of scalar multiplication:
		\begin{equation}
			1v=v, 
		\end{equation}
		where $1$ denotes the multiplicative identity in $F$.
	\end{itemize}
	An abelian group for which there is a scalar multiplication map called a vector space over field $F$. The elements of $F$ are called scalars.
	
	\section{Full definition of module}
	\label{prop_mod}
	Let $M$ be be an abelian group under the operation $+$ and let $R$ be a ring. The map 
	\begin{equation}
		R\times M\to M:\quad (\alpha,x) \to \alpha x,
	\end{equation}
	such that for all $x,y\in M$ and all $\alpha,\beta\in R$ it has the properties:
	\begin{itemize}
		\item Distributivity of scalar multiplication with respect to module addition:
		\begin{equation}
			\alpha(x+y)=\alpha x+\alpha y.
		\end{equation}
		\item Distributivity of scalar multiplication with respect to ring addition:
		\begin{equation}
			(\alpha+\beta)x=\alpha x + \beta x.
		\end{equation}
		\item Compatibility of scalar multiplication with ring multiplication:
		\begin{equation}
			\alpha(\beta x)=(\alpha\beta)x.
		\end{equation}
		\item Identity element of scalar multiplication:
		\begin{equation}
			1v=v, 
		\end{equation}
		where $1$ denotes the multiplicative identity in $R$.
	\end{itemize}
	is called the scalar multiplication on module, which has the properties as same as the forms of scalar multiplication on vector space. An abelian group for which there is a scalar multiplication on module is called a left-$R$ module. And the right-$R$ module is defined similarly in terms of a map: $M\times R\to M$. If $R$ is commutative, then left $R$-modules are the same as right $R$-modules and are simply called $R$-modules. If $R$ is a field, then $R$-modules are vector spaces.

	\section{Full definition of inner product on vector space}
	\label{prop_inner}
	An inner product on a vector space $V$ over field $F$ is a map:
	\begin{equation}
		V\times V\to F: \langle x,y\rangle\to m,
	\end{equation}
	that satisfies properties as follows for all vectors $x,y,z\in Z$ and all scalars $a,b\in F$:
	\begin{itemize}
		\item Conjugate symmetry:
		\begin{equation}
			\langle x,y\rangle=\overline{\langle y,x\rangle}
		\end{equation}
		\item Linearity in the first argument:
		\begin{equation}
			\langle ax+by,z\rangle=a\langle x,z\rangle+b \langle y,z \rangle
		\end{equation}
		\item Positive-definiteness: if $x$ is not zero, then:
		\begin{equation}
			\langle x,x\rangle>0
		\end{equation}
	\end{itemize}
	
	\section{Best Hyperparameters Settings}
	\label{hyper_para}
	The best hyperparameter settings for ModulE$_\mathbb{H,H}$ is shown in Table \ref{para-table}.
	\begin{table}[h]
		\small
		\caption{Best hyperparameters settings.}
		\label{para-table}
		\vskip 0.15in
		\begin{center}
			\begin{tabular}{ccccccccc}
				\toprule
				\textbf{Dataset} & Epoch & Batch Size & $k$ & $p$ & $\lambda$ & $\lambda_1$ & $\lambda_2$ & $\lambda_3$\\\midrule
				FB15k-237 & 200 & 300 & 128 & 3 & 0.045 & 2.0 & 0.5 & 2.0 \\
				WN18RR & 200 & 500 & 128 & 3 & 0.08 & 2.0 & 0.5 & 2.0 \\
				YAGO3-10 & 200 & 1000 & 128 & 3 & 0.005 & 2.0 & 0.5 & 2.0 \\
				\bottomrule
			\end{tabular}
		\end{center}
		\vskip -0.1in
	\end{table}
	
\end{document}